\documentclass[11pt]{article}
\usepackage[utf8]{inputenc}
\usepackage[T1]{fontenc}
\input{glyphtounicode}
\usepackage{lmodern}
\usepackage[margin=1in]{geometry}
\usepackage{microtype}
\usepackage{amsmath,amssymb}
\usepackage{graphicx}
\usepackage{booktabs}
\usepackage{tabularx}
\usepackage{array}
\usepackage{caption}
\usepackage{fancyvrb}
\usepackage{placeins}
\usepackage[none]{hyphenat}
\usepackage{xurl}
\usepackage{cite}
\usepackage[colorlinks=true,linkcolor=blue,citecolor=blue,urlcolor=blue]{hyperref}
\graphicspath{{./}{figures/}}
\title{What Training Data Teaches RL Memory Agents: An Empirical Study of Curriculum Effects in Memory-Augmented QA}
\author{
Xinjie He$^{1}$, Zhiyuan Lin$^{2}$, Su Liu$^{2}$, Jialun Wu$^{3}$,\\
Qiyang Xie$^{4}$, Weikai Zhou$^{2}$, and Shuai Xiao$^{2}$\\[0.5em]
$^{1}$Columbia University \\
$^{2}$Independent Researcher \\
$^{3}$Johns Hopkins University \\
$^{4}$Northeastern University
}
\date{May 2026}
\begin{document}
\maketitle
\begin{abstract}
Reinforcement learning (RL) has emerged as a viable recipe for training LLM agents to reason over external memory banks in multi-session dialogue. Existing work trains exclusively on a single benchmark, leaving open how the composition of training data shapes the skills a memory agent acquires. We present a controlled empirical study that holds architecture, RL algorithm, and all hyperparameters fixed and varies only the training curriculum across three conditions: in-domain (LoCoMo), mixed-benchmark (LoCoMo + LongMemEval), and out-of-domain (LongMemEval only). Across two benchmarks and ten question types, curriculum composition acts as a fine-grained lever on specialization rather than a uniform scaling factor on performance. The mixed curriculum yields the strongest overall F1 on both evaluation sets. Training on a narrow out-of-domain set transfers a targeted skill --- temporal reasoning --- despite weak aggregate performance. And per-type differences substantially exceed aggregate differences, indicating that single-number benchmark comparisons systematically underreport curriculum effects. We further report two practical lessons from adapting GRPO to a single-GPU regime: cross-benchmark mixing requires filtering format-specific noise from memory banks to preserve training signal, and binary exact-match reward produces no learning signal at the small group sizes ($G=4$) required on one GPU, motivating continuous reward functions in this regime. 
\end{abstract}
\section{Introduction}
Large language models operate within fixed context windows, with no persistent memory across interactions. This limitation is acute in multi-session dialogue, where users expect the system to recall preferences, events, and relationships from prior conversations.
Recent work addresses this by augmenting LLMs with external memory banks --- structured stores that persist across sessions and support retrieval at inference time \cite{mem0,amem,langmem,zep}. A key challenge is learning to use retrieved memories well: selecting relevant entries from a noisy candidate set, reasoning across them, and producing concise answers. Heuristic pipelines rely on fixed retrieval rules; RL-based approaches \cite{memoryr1} instead let the agent discover such selection and reasoning patterns through outcome-driven training, achieving competitive results with small supervision budgets.
The framing that has dominated this line of work is training on a single benchmark and reporting aggregate scores. Yet memory-augmented QA is a multi-skill problem: different question types exercise different combinations of retrieval precision, multi-hop composition, temporal ordering, and knowledge-update tracking. When a single benchmark exercises only a subset of these skills, single-benchmark training implicitly selects which capabilities the RL signal reinforces. This raises two questions the existing literature leaves open. First, does exposing the RL signal to a broader source of question types --- mixing benchmarks --- bias the policy toward a more general memory skill set, or does it dilute the in-domain skill without cross-benchmark gains? Second, when only a small out-of-domain set is available, does RL surface a targeted capability, or does it fail entirely? We refer to the resulting per-question-type profile as the policy's \emph{specialization}: systematic variation in per-type strengths induced by training curriculum composition.
To answer these, we fix architecture (Qwen-2.5-7B with LoRA), RL algorithm (GRPO), and all optimization hyperparameters, and vary only the training curriculum. Config A replicates the single-benchmark baseline of prior RL-memory work with 152 LoCoMo QA pairs. Config B mixes LoCoMo with 60 LongMemEval pairs (212 total). Config C trains on the 60 LongMemEval pairs alone. All three are evaluated on both LoCoMo (1,307 test questions, 4 types) and LongMemEval (415 test questions, 6 types).
Our contributions are as follows. (i) We present a controlled curriculum study for RL-based memory agents in which architecture, algorithm, and hyperparameters are held fixed and only the training source varies. Under this design, per-question-type differences substantially exceed aggregate differences, so curriculum composition acts as a lever on specialization rather than on overall accuracy (Section 4.2). (ii) We draw practical guidance from this design space: mixing benchmarks yields the strongest generalist in our setting, a narrow out-of-domain set can induce a targeted behavior (temporal reasoning), and the transition between targeted specialization and more stable aggregate gains appears between roughly 60 and 150 training examples in our setup (Section 5.3). (iii) We document two engineering findings that affect the reproducibility of GRPO-based memory training on a single GPU: cross-benchmark data requires filtering format-specific noise from the memory bank (Section 5.1), and binary exact-match reward collapses to zero advantage at $G=4$, strongly motivating continuous reward functions when the large-group regime is not available (Section 5.2).
\begin{figure}[htbp]
  \centering
  \includegraphics[width=0.95\linewidth]{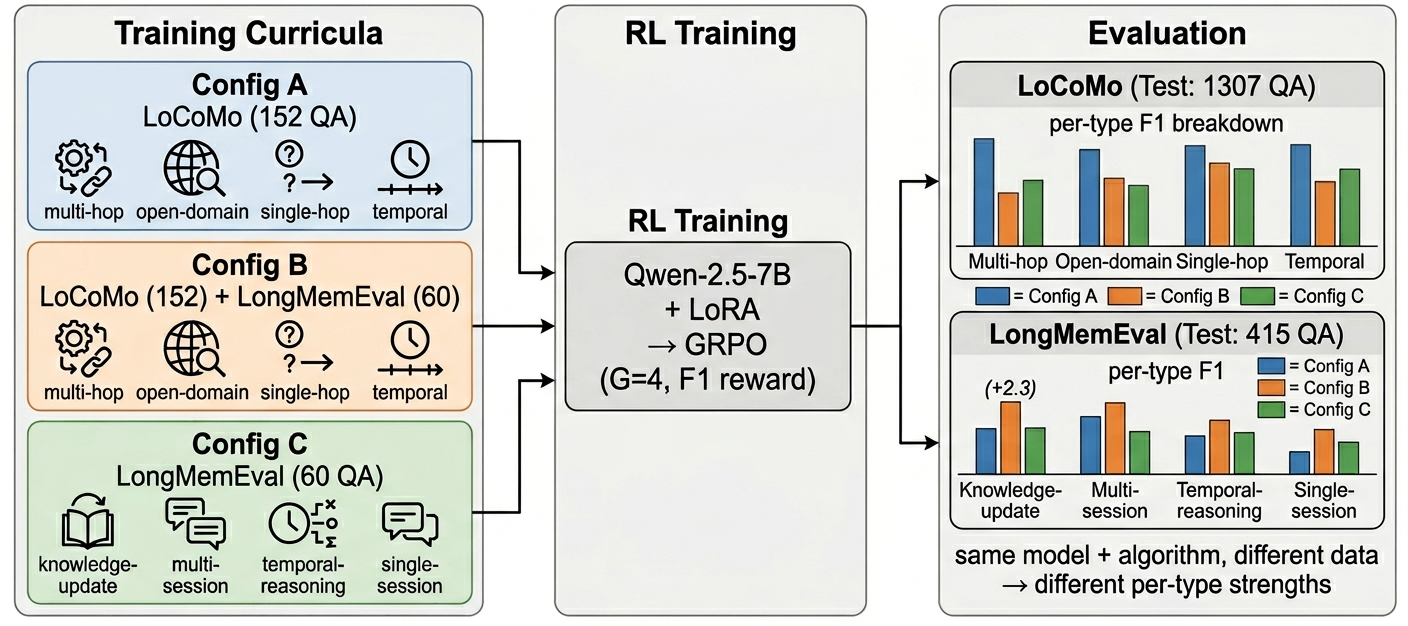}
  \caption{Experimental design. Three curricula, identical training recipe, evaluated on two benchmarks with per-type F1 breakdown.}
\end{figure}

\FloatBarrier
\section{Related Work}
\subsection{Memory-Augmented LLM Agents}
The challenge of equipping LLMs with persistent memory has motivated several architectural approaches. Early agent frameworks such as Reflexion \cite{reflexion} demonstrated the value of persistent state for multi-step reasoning, though their memory policies are largely handcrafted. Recent memory systems build on this foundation: MemGPT \cite{memgpt} treats the LLM's context window as a virtual memory with OS-inspired paging between a primary and secondary store. Mem0 \cite{mem0} provides a modular memory system with explicit CRUD operations. A-Mem \cite{amem} introduces dynamic agentic memory with structured entries. LangMem \cite{langmem} chains memory entries across sessions. Zep \cite{zep} employs a temporally-aware knowledge graph for agent memory, benchmarking directly against MemGPT on Deep Memory Retrieval and LongMemEval. These systems rely on heuristic memory management policies.
Recent work has begun applying RL to memory-augmented agents. Memory-R1 \cite{memoryr1} trains both a Memory Manager (for CRUD operations) and an Answer Agent (for memory-grounded QA) using GRPO \cite{deepseekmath}, achieving strong results with 152 training examples. Our work focuses on the Answer Agent component --- the agent that selects and reasons over retrieved memories to answer questions --- and studies how training data composition affects its learned skills. We use heuristic memory construction and focus our RL training on the answer generation policy, isolating the effect of curriculum composition from memory management quality.
\subsection{Benchmarks for Long-Term Memory}
LoCoMo \cite{locomo} (ACL 2024) provides multi-session dialogues averaging 26,000 tokens with approximately 200 questions per dialogue spanning single-hop, multi-hop, open-domain, and temporal reasoning. It features rich multi-party conversations between named speakers.
LongMemEval \cite{longmemeval} (ICLR 2025) offers 500 questions across six categories: single-session-user, single-session-assistant, single-session-preference, multi-session, temporal-reasoning, and knowledge-update. Each question is paired with approximately 40 haystack sessions in user-assistant chat format, testing precise retrieval from large conversation histories. Both benchmarks exercise retrieval-augmented generation \cite{rag} over dialogue, relying on dense retrievers \cite{dpr,sbert} to surface relevant context.
These benchmarks test complementary skills: LoCoMo emphasizes reasoning over rich dialogues, while LongMemEval emphasizes precise retrieval and temporal reasoning. This complementarity motivates our mixed-curriculum approach.
\subsection{Curriculum Learning for RL-Based LLM Training}
Curriculum learning --- structuring training data to improve learning --- is a long-standing idea \cite{curriculumlearning,automatedcurriculum} that has been revisited for LLM post-training. Recent work in this area has focused on the \emph{difficulty axis}: difficulty-based curricula \cite{easyhard,vcrl} schedule examples from easy to hard, and distribution-level curricula \cite{dump} reweight data sources to manage effective problem difficulty. In RL specifically, reward variance and reward shaping have both been used as proxies for difficulty.
Our axis is different. We study \textbf{source composition} --- which benchmarks the training data is drawn from --- while keeping the per-example difficulty signal unchanged. We do not order examples, reweight them, or prune them by difficulty; Configs A, B, and C see the same per-example reward function under the same optimizer. The question is whether widening the source distribution at fixed per-example signal changes which skills the policy acquires. Under the tightly controlled design we adopt, this axis is orthogonal to difficulty-based curricula, and the two could be combined in future work. To our knowledge, source-level curriculum composition has not been studied in the RL-for-memory-agents setting.
\subsection{GRPO and Reward Design}
Group Relative Policy Optimization (GRPO) \cite{deepseekmath} computes advantages relative to a group of $G$ sampled completions, eliminating the need for a learned value function used in PPO-based RLHF \cite{instructgpt,ppo}. GRPO also sidesteps the preference-pair formulation of DPO \cite{dpo} by scoring completions directly against a task reward. It was used successfully in DeepSeek-R1 \cite{deepseekr1} and subsequent RL-for-LLM work \cite{memoryr1}.
A less-studied aspect of GRPO is the interaction between reward sparsity and group size. We show that binary exact-match reward, which is used for the Memory-R1 Answer Agent \cite{memoryr1} and for DeepSeek-R1 verifiable tasks \cite{deepseekr1} in the large-group regime, collapses to zero within-group variance at the small group sizes ($G=4$) that fit on a single GPU, producing no task-relevant gradient for the Answer Agent. We treat this as a practical barrier to single-GPU reproduction rather than a theoretical claim, and analyze it in Section 5.2.
Taken together, these threads --- memory-augmented agents \cite{mem0,amem,langmem,zep,memgpt}, RL for memory \cite{memoryr1}, long-term dialogue benchmarks \cite{locomo,longmemeval}, and curriculum learning \cite{easyhard,vcrl,dump} --- leave this intersection comparatively underexplored. Prior RL-for-memory work fixes a single source benchmark, so architecture/algorithm effects cannot be separated from data-composition effects. Prior curriculum work targets general mathematical or code reasoning rather than memory-grounded QA. And prior GRPO recipes assume a large-group regime in which the reward-variance issue is hidden. Our work fills this gap with a controlled source-level comparison and reports the small-group reward-variance constraint as a practical consequence.
\section{Method}
\subsection{Task: Memory-Augmented Question Answering}
Given a multi-session dialogue history and a question, the task is to answer the question using information distributed across sessions. The agent receives a set of retrieved memory entries (extracted from the dialogue) and must select relevant entries, reason over them, and produce a concise answer.
Formally, the agent policy $\pi_\theta$ maps a question $q$ and retrieved memories $\mathcal{M}_{\text{ret}}$ to an answer $y$:
$$y \sim \pi_\theta(\cdot \mid q, \mathcal{M}_{\text{ret}})$$
where $\mathcal{M}_{\text{ret}}$ is a set of top-$k$ memory entries retrieved via embedding similarity from the full memory bank.
\subsection{Training Configurations}
All three configurations share identical hyperparameters, differing only in training data:
\begin{table}[htbp]
  \centering
  \small
  \caption{Training configurations. Data composition is the only variable across runs.}
  \begin{tabularx}{\linewidth}{>{\raggedright\arraybackslash}p{0.23\linewidth} >{\raggedright\arraybackslash}p{0.29\linewidth} >{\centering\arraybackslash}p{0.12\linewidth} >{\raggedright\arraybackslash}X}
    \toprule
    Config & Training Data & QA Pairs & Source Benchmarks \\
    \midrule
    A (single-benchmark) & LoCoMo only & 152 & LoCoMo \\
    B (mixed) & LoCoMo + LongMemEval & 212 & LoCoMo + LongMemEval \\
    C (specialist) & LongMemEval only & 60 & LongMemEval \\
    \bottomrule
  \end{tabularx}
\end{table}
\subsection{RL Training with GRPO}
We fine-tune Qwen-2.5-7B-Instruct \cite{qwen25} with LoRA \cite{lora} ($r=16$, $\alpha=32$), a parameter-efficient adapter approach \cite{adapters} that keeps most of the base model frozen. Training uses GRPO \cite{deepseekmath} with group size $G=4$, the largest value that fits on a single 48 GB GPU. The reward is the sum of a token-level F1 term between extracted answer and gold answer (primary) and a small XML format term capped at 0.2 (secondary); we initially adopted binary exact-match reward following prior work \cite{memoryr1}, but this produced zero gradient signal at $G=4$ and the switch to F1 is analyzed in Section 5.2. We use a learning rate of $5 \times 10^{-6}$ with cosine decay, effective batch size 4 (batch size 1 with 4 gradient accumulation steps), and a 512-token completion cap. Configs A and B train for 2 epochs and Config C for 3, corresponding to roughly 76, 106, and 45 gradient steps respectively; the larger epoch count on C partially compensates for its smaller dataset but the total number of updates remains unequal. All training runs on a single NVIDIA L40S (48 GB).
The agent is prompted to output structured XML with \texttt{<selected\_memories>}, \texttt{<reasoning>}, and \texttt{<answer>} tags. The format reward provides a small bonus for correct structure while the F1 reward drives task learning. Full hyperparameters are in Appendix A.
\subsection{Memory Construction and Retrieval}
For each example we construct a memory bank from dialogue sessions. We first extract substantive turns, keeping only those longer than five words and discarding greetings. For chat-format data (LongMemEval), we additionally filter out long assistant responses (more than 40 words), retaining user turns and short confirmations; this reduces the bank by roughly 50\% and improves retrieval precision, as analyzed in Section 5.1. The memory bank is capped at 200 entries per conversation. At inference time we retrieve the top 60 entries by embedding similarity using a Sentence-BERT-style encoder \cite{sbert} (all-MiniLM-L6-v2), and truncate any prompt exceeding roughly 3,000 tokens to prevent context overflow.
\subsection{Evaluation}
We evaluate on two benchmarks. LoCoMo \cite{locomo} contributes 1,307 test questions drawn from 10 dialogues and split across four question types: single-hop (238), multi-hop (279), open-domain (705), and temporal (85). LongMemEval \cite{longmemeval} contributes 415 test questions across 415 conversations in six categories: single-session-user (55), single-session-assistant (39), single-session-preference (20), multi-session (118), temporal-reasoning (119), and knowledge-update (64).
Token-level F1 after answer normalization is our primary metric. We additionally collect an LLM-as-Judge rating on a 1--5 scale using Claude 3 Haiku to score each answer on accuracy, relevance, and completeness, which we use as a compatibility check against a coarser semantic evaluator (Section 4.3). Throughout, we use training progress bins Q1--Q4 to denote the four equal-length quartiles of gradient steps within a run.
\FloatBarrier
\section{Results}
\subsection{Main Results}
Table 2 presents overall results. Config B (mixed curriculum) achieves the highest F1 on both benchmarks, outperforming the baseline by +0.012 on LoCoMo and +0.014 on LongMemEval.
\begin{table}[htbp]
  \centering
  \small
  \caption{Overall F1 scores. Best RL-trained configuration in bold.}
  \begin{tabularx}{\linewidth}{>{\raggedright\arraybackslash}p{0.28\linewidth} >{\raggedright\arraybackslash}p{0.24\linewidth} >{\centering\arraybackslash}p{0.16\linewidth} >{\centering\arraybackslash}X}
    \toprule
    Model & Training Data & LoCoMo F1 & LongMemEval F1 \\
    \midrule
    Baseline (no RL) & --- & 0.119 & 0.141 \\
    Config A & LoCoMo (152) & 0.123 & 0.147 \\
    Config B & Mixed (212) & \textbf{0.131} & \textbf{0.155} \\
    Config C & LongMemEval (60) & 0.120 & 0.151 \\
    \bottomrule
  \end{tabularx}
\end{table}
While overall gains are modest, the per-question-type analysis reveals substantially larger and more informative differences.
\subsection{Per-Question-Type Analysis: Where Curriculum Composition Matters}
The central empirical finding of this paper is that different training curricula produce models whose per-type profiles diverge far more than their aggregate scores suggest. Tables 3 and 4 show per-type differences that are, in magnitude, several times larger than the gap in overall F1 between any two configurations.
Different question types stress different combinations of retrieval filtering and compositional reasoning. Mixed curricula expose the policy to a broader distribution of such behaviors, while narrow curricula reinforce a smaller subset. We therefore expect curriculum composition to shape specialization patterns more strongly than aggregate performance. The per-type results below are broadly consistent with this interpretation.
\begin{table}[htbp]
  \centering
  \small
  \caption{LoCoMo F1 by question type. Bold indicates the best-performing RL-trained configuration; Best $\Delta$ is that configuration's gain over baseline, or "---" if no trained configuration exceeds baseline.}
  \begin{tabularx}{\linewidth}{>{\raggedright\arraybackslash}p{0.26\linewidth} >{\centering\arraybackslash}p{0.06\linewidth} >{\centering\arraybackslash}p{0.10\linewidth} >{\centering\arraybackslash}p{0.10\linewidth} >{\centering\arraybackslash}p{0.10\linewidth} >{\centering\arraybackslash}p{0.10\linewidth} >{\centering\arraybackslash}X}
    \toprule
    Question Type & n & Baseline & Config A & Config B & Config C & Best $\Delta$ \\
    \midrule
    multi-hop & 279 & 0.080 & \textbf{0.093} & 0.083 & 0.080 & +0.013 \\
    open-domain & 705 & 0.139 & \textbf{0.145} & 0.132 & 0.139 & +0.006 \\
    single-hop & 238 & 0.129 & \textbf{0.132} & 0.128 & 0.129 & +0.003 \\
    temporal & 85 & 0.116 & \textbf{0.131} & 0.121 & 0.116 & +0.015 \\
    \bottomrule
  \end{tabularx}
\end{table}
On LoCoMo, per-category F1 gaps between configurations are small --- within roughly 0.02 F1 in every row of Table 3. Config A shows the largest per-category gains over the no-RL baseline, most visibly on temporal (+0.015) and multi-hop (+0.013), reflecting the advantage of in-domain training on category-specific retrieval patterns.
We note that Config B has 40\% more training examples than Config A (212 vs 152). However, the additional 60 examples come from LongMemEval, and Config C shows that 60 LongMemEval examples alone produce negligible overall improvement. This suggests Config B's gains stem from curriculum diversity --- the combination of LoCoMo and LongMemEval skills --- rather than simply having more data.
\begin{table}[htbp]
  \centering
  \small
  \caption{LongMemEval F1 by question type. Bold indicates the best-performing RL-trained configuration; Best $\Delta$ is that configuration's gain over baseline, or "---" if no trained configuration exceeds baseline.}
  \begin{tabularx}{\linewidth}{>{\raggedright\arraybackslash}p{0.26\linewidth} >{\centering\arraybackslash}p{0.06\linewidth} >{\centering\arraybackslash}p{0.10\linewidth} >{\centering\arraybackslash}p{0.10\linewidth} >{\centering\arraybackslash}p{0.10\linewidth} >{\centering\arraybackslash}p{0.10\linewidth} >{\centering\arraybackslash}X}
    \toprule
    Question Type & n & Baseline & Config A & Config B & Config C & Best $\Delta$ \\
    \midrule
    knowledge-update & 64 & 0.124 & 0.121 & \textbf{0.147} & 0.123 & +0.023 \\
    multi-session & 118 & 0.073 & 0.078 & \textbf{0.083} & 0.079 & +0.010 \\
    single-session-assistant & 39 & 0.106 & \textbf{0.163} & 0.159 & 0.149 & +0.057 \\
    single-session-preference & 20 & 0.190 & 0.169 & 0.169 & \textbf{0.177} & --- \\
    single-session-user & 55 & 0.196 & 0.229 & \textbf{0.231} & 0.221 & +0.035 \\
    temporal-reasoning & 119 & 0.194 & 0.182 & 0.192 & \textbf{0.202} & +0.008 \\
    \bottomrule
  \end{tabularx}
\end{table}
We read Tables 3 and 4 as a set of directional observations; we do not claim individual per-type gains are statistically robust at the evaluation sizes available (category $n$ ranges from 20 to 705).
Different curricula concentrate their benefit on different question types. Config A's largest LoCoMo gains (temporal +0.015, multi-hop +0.013) target categories that stress in-domain retrieval and reasoning. Config B's largest LongMemEval gains (knowledge-update +0.023, single-session-user +0.035) target categories that stress cross-session fact tracking and preference retrieval. Config B's aggregate advantage on both benchmarks (Table 2) is therefore not a uniform lift but a redistribution across categories. The small out-of-domain curriculum produces a more concentrated specialization: Config C posts the highest temporal-reasoning score (0.202) despite only a modest aggregate lift over the no-RL baseline, suggesting that 60 LongMemEval examples suffice to expose the policy to temporal composition patterns but not to broadly strengthen retrieval. On categories where no trained model exceeds baseline --- notably single-session-preference ($n=20$) --- the differences are small in absolute terms and likely reflect sampling variation at that category size. The largest single per-type gain in the study (+0.057 on single-session-assistant, Config A, $n=39$) should be read cautiously for the same reason. The aggregate-vs-per-type gap --- per-type differences several times the size of any overall F1 gap --- is the pattern we emphasize, because it is present across categories of different sizes and across both benchmarks.
\begin{figure}[htbp]
  \centering
  \includegraphics[width=0.95\linewidth]{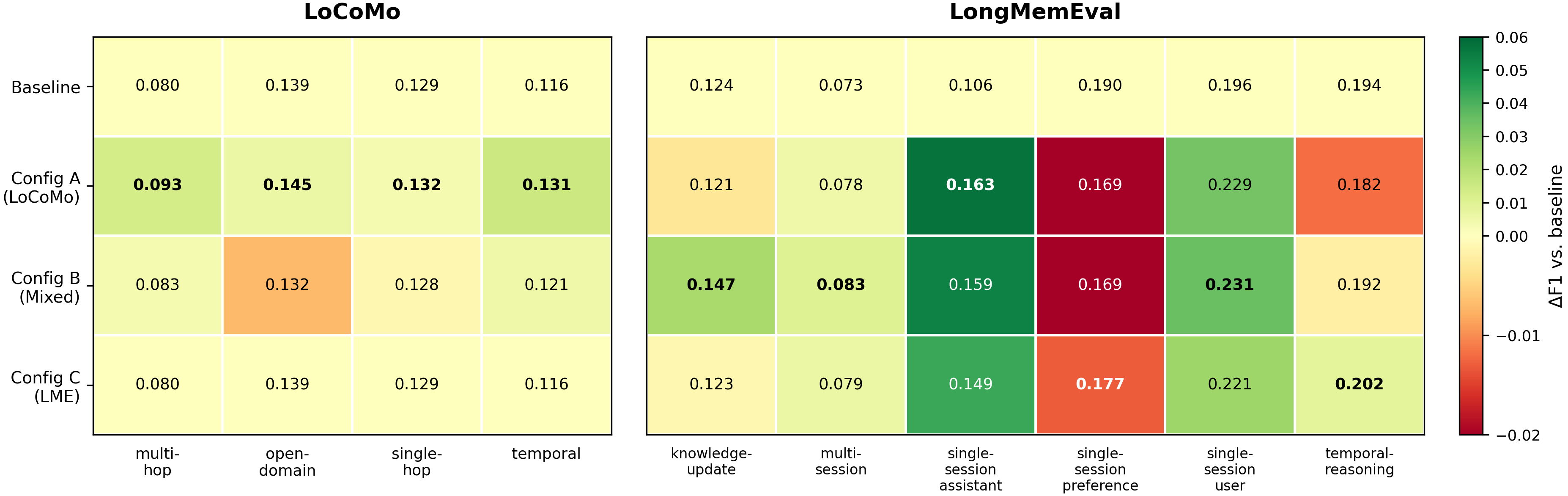}
  \caption{Per-question-type F1 across all models and both benchmarks. Color intensity shows delta from baseline (green = improvement, red = regression). Bold values indicate the best RL-trained configuration per type. Curriculum effects concentrate in specific question types, with per-type differences several times larger than overall gaps.}
\end{figure}
\subsection{LLM-as-Judge as a Compatibility Check}
A natural concern about using token-level F1 as the primary metric is that F1 under-rewards semantically correct but lexically different answers, which could in principle hide curriculum effects that a semantic judge would surface. To check for this, we score every model's output with Claude 3 Haiku on a 1--5 scale covering accuracy, relevance, and completeness (Appendix C). Mean scores cluster tightly across all four models (3.22--3.39), and no model falls below a mean of 3.0 on either benchmark, so RL training does not degrade answer quality in a way the judge would flag. Critically, the per-model ordering under the judge is not systematically different from the ordering under F1 --- the judge does not reveal a hidden curriculum effect that F1 missed. We therefore keep F1 as the primary metric for the rest of the analysis; we do not read the judge as a competing signal, only as a compatibility check against a coarser semantic evaluator.
\FloatBarrier
\section{Analysis}
\subsection{Memory Bank Preprocessing Matters for Cross-Benchmark Transfer}
LongMemEval uses a user-assistant chat format where assistant responses are typically long, generic advice (e.g., "That sounds great! Here are some tips for..."). When included in the memory bank, these responses constitute approximately 50\% of entries but contribute no useful facts for answering questions about the user.
We compared two versions of the Config B training run. The first version (v1) stored all dialogue turns as memories, yielding about 466 entries per example with roughly 50\% assistant filler. The second version (v2) filtered out long assistant responses, leaving about 231 user-focused entries per example.
\begin{table}[htbp]
  \centering
  \small
  \caption{Effect of memory bank preprocessing on Config B training signal.}
  \begin{tabularx}{\linewidth}{>{\raggedright\arraybackslash}p{0.24\linewidth} >{\raggedright\arraybackslash}p{0.24\linewidth} >{\raggedright\arraybackslash}X}
    \toprule
    Config B & Training F1 mean & Q1$\rightarrow$Q4 reward trend \\
    \midrule
    v1 (unfiltered) & 0.159 & 0.324 $\rightarrow$ 0.364 \\
    v2 (filtered) & 0.194 (+22\%) & 0.348 $\rightarrow$ 0.403 \\
    \bottomrule
  \end{tabularx}
\end{table}
The filtered version showed 22\% higher mean F1 during training and a stronger upward trend. This demonstrates that memory bank quality directly affects RL training signal quality. Practitioners building cross-benchmark curricula should preprocess data to remove format-specific noise rather than naively mixing sources.
\subsection{Reward Sparsity and GRPO Group Size}
Binary exact-match (EM) reward assigns 1.0 for exact matches and 0.0 otherwise. GRPO standardizes rewards within a group of $G$ candidates: $A_i = (r_i - \bar{r}) / \sigma_r$. When all candidates in a group receive the same reward, $\sigma_r = 0$ and no gradient flows. With large group sizes ($G \approx 128$) some candidates are likely to hit exact match, preserving within-group variance; with $G=4$ on a single GPU, the probability of any candidate exactly matching the gold answer is near zero for open-ended QA.
In our initial experiments with EM reward, the task reward component was 0.0 on every training step across the full run (several hundred steps). The total reward was capped at 0.2 (the format reward ceiling), and GRPO produced no task-relevant gradient signal. Switching to token-level F1 reward resolved this immediately: at step 5, task reward was 0.23 with std=0.03, providing sufficient variance for learning.
We do not claim this issue is unique to GRPO --- sparse rewards are a known difficulty. Rather, GRPO makes the variance collapse especially visible because advantages are normalized within each sampled group. Under continuous reward (F1), within-group variance is preserved at any group size; under binary reward in the small-group regime it is not. The practical rule of thumb is straightforward: on single-GPU GRPO, prefer continuous reward functions. This also intersects with known concerns about reward specification, where overly sparse or mis-specified rewards cause agents to exploit proxy signals rather than learn the intended skill \cite{rewardhacking}.
\subsection{Training Set Size and Observed Thresholds}
Config C's mean reward declines from Q1 to Q4 ($0.344 \to 0.325$) across its three epochs, while Configs A and B both trend upward. This is consistent with Config C revisiting each example three times and beginning to fit the training distribution rather than continuing to improve on held-out questions.
At the same time, Config C is not simply noise around the no-RL baseline: it achieves the highest temporal-reasoning score of any configuration (0.202), indicating that a narrow 60-example set can transfer a targeted capability under GRPO. Combined with the upward trajectories of Configs A ($n=152$) and B ($n=212$), the pattern in our study is that very small training sets ($n \approx 60$) can induce targeted behaviors but do not yield reliable aggregate improvement, while roughly 150 examples suffices for positive aggregate deltas. The transition between targeted specialization and more stable aggregate gains therefore appears between roughly 60 and 150 examples in our setup; a finer scan would be needed to pin it down.
\begin{table}[htbp]
  \centering
  \small
  \caption{Reward trajectory and overall F1 gain by training set size. F1 gain is the absolute change over baseline; the range spans LoCoMo and LongMemEval test sets.}
  \begin{tabularx}{\linewidth}{>{\centering\arraybackslash}p{0.10\linewidth} >{\centering\arraybackslash}p{0.12\linewidth} >{\centering\arraybackslash}p{0.10\linewidth} >{\raggedright\arraybackslash}p{0.24\linewidth} >{\centering\arraybackslash}X}
    \toprule
    Config & Examples & Epochs & Reward trend & Overall F1 gain \\
    \midrule
    A & 152 & 2 & $\uparrow$ (0.298 $\rightarrow$ 0.351) & +0.004 to +0.006 \\
    B & 212 & 2 & $\uparrow$ (0.348 $\rightarrow$ 0.403) & +0.012 to +0.014 \\
    C & 60 & 3 & $\downarrow$ (0.344 $\rightarrow$ 0.325) & +0.001 to +0.010 \\
    \bottomrule
  \end{tabularx}
\end{table}
\FloatBarrier
\section{Limitations and Future Work}
\textbf{Single-GPU constraints.} Our experiments use LoRA \cite{lora} with $G=4$, compared to full fine-tuning with $G \approx 128$ in prior RL-for-memory work \cite{memoryr1}. Memory-efficient variants such as QLoRA \cite{qlora} could further reduce the footprint but were not needed on the L40S. The smaller group size necessitated switching from exact-match to F1 reward and likely limits absolute performance. Full fine-tuning, and the resulting larger groups, would strengthen the curriculum comparison.
\textbf{Heuristic memory construction.} We use heuristic extraction rather than a trained Memory Manager. Prior work \cite{memoryr1} attributes roughly 7.5 F1 points to RL-trained memory management. Whether curriculum composition has analogous specialization effects on a learned Memory Manager is an open question.
\textbf{Limited benchmarks.} We evaluate on LoCoMo and LongMemEval. MSC \cite{msc} and EverMemBench would broaden the evaluation axis.
\textbf{Single model family.} All experiments use Qwen-2.5-7B-Instruct. Replication across Llama-family and Mistral-family backbones would test whether the specialization pattern we report is model-specific.
\textbf{Coarser data-size scan.} Our three configurations test 60, 152, and 212 training examples. A finer scan between 60 and 152 would sharpen any claim about the effective data-size regime.
\section{Conclusion}
We presented a controlled study of training data composition in RL-based memory agents, holding model, algorithm, and optimizer fixed and varying only the curriculum source. Across two benchmarks and ten question types, curriculum composition acts more like a lever on specialization than a uniform scaling factor on aggregate performance: the mixed curriculum produces the strongest generalist in our setting, a narrow out-of-domain set transfers a specific temporal-reasoning behavior despite weak aggregate scores, and per-type gaps are several times the size of any aggregate gap. Alongside the curriculum findings, we report two practical lessons from adapting GRPO to a single GPU: cross-benchmark data must be preprocessed to remove format-specific noise to preserve training signal, and binary exact-match reward collapses to zero advantage at the small group sizes that fit on one GPU, strongly motivating continuous reward functions in this regime. We hope this controlled slice of the design space is useful to practitioners reasoning about how to allocate limited RL training budget across benchmarks.
\section{Reproducibility Statement}
All experiments were conducted on a single NVIDIA L40S GPU (48 GB) using LoRA fine-tuning. Training configurations, hyperparameters, and evaluation scripts are provided in the supplementary material.

Trained checkpoints and prediction files are available at:\\
\url{https://github.com/EvaxHe/rl-memory-curriculum}.\\
The exact code revision used for this preprint is tagged \texttt{v1.0-arxiv} in that repository.
\bibliographystyle{unsrt}
\bibliography{references}
\appendix
\section{Hyperparameters}
\begin{table}[htbp]
  \centering
  \small
  \caption{Full training hyperparameters.}
  \begin{tabular}{>{\raggedright\arraybackslash}p{0.40\linewidth} >{\raggedright\arraybackslash}p{0.52\linewidth}}
    \toprule
    Parameter & Value \\
    \midrule
    Model & Qwen-2.5-7B-Instruct \\
    LoRA rank ($r$) & 16 \\
    LoRA alpha ($\alpha$) & 32 \\
    LoRA target modules & q\_proj, k\_proj, v\_proj, o\_proj, up\_proj, down\_proj, gate\_proj \\
    LoRA dropout & 0.05 \\
    GRPO group size ($G$) & 4 \\
    Batch size & 1 \\
    Gradient accumulation & 4 \\
    Effective batch size & 4 \\
    Learning rate & $5 \times 10^{-6}$ \\
    LR schedule & Cosine with 10\% warmup \\
    Weight decay & 0.1 \\
    Adam $\beta_1, \beta_2$ & 0.9, 0.99 \\
    Max grad norm & 1.0 \\
    Max completion length & 512 \\
    Retrieval top-$k$ & 60 \\
    Max memories per conversation & 200 \\
    Embedding model & all-MiniLM-L6-v2 \\
    Precision & bf16 \\
    Hardware & 1$\times$ NVIDIA L40S 48 GB \\
    \bottomrule
  \end{tabular}
\end{table}
\FloatBarrier
\section{Prompt Template}
\begin{scriptsize}
\begin{verbatim}
You are an Answer Agent for a conversational AI assistant.
You have access to a memory bank containing facts from past conversations.

Given a question and retrieved memories, you must:
1. Select the most relevant memories for answering the question.
2. Reason step-by-step using the selected memories.
3. Provide a concise, accurate answer.

Output format:
<selected_memories>[list the memory IDs or snippets]</selected_memories>
<reasoning>[your step-by-step reasoning]</reasoning>
<answer>[your final answer - be concise]</answer>
\end{verbatim}
\end{scriptsize}
\section{LLM-as-Judge Scores}
\begin{table}[htbp]
  \centering
  \small
  \caption{Mean LLM-as-Judge scores (Claude 3 Haiku, 1--5 scale).}
  \begin{tabularx}{\linewidth}{>{\raggedright\arraybackslash}X >{\centering\arraybackslash}X >{\centering\arraybackslash}X}
    \toprule
    Model & LoCoMo & LongMemEval \\
    \midrule
    Baseline & 3.27 & 3.39 \\
    Config A & 3.27 & 3.30 \\
    Config B & 3.22 & 3.36 \\
    Config C & 3.30 & 3.39 \\
    \bottomrule
  \end{tabularx}
\end{table}
See Section 4.3 for interpretation.
\FloatBarrier
\section{Qualitative Examples}
\paragraph{Example 1: Knowledge-update (Config B correct, Baseline wrong)}
Question: "What is the name of the music streaming service I have been using lately?" Gold answer: "Spotify"
\begin{table}[htbp]
  \centering
  \small
  \caption{Model answers for Example 1: Knowledge-update (Config B correct, Baseline wrong).}
  \begin{tabularx}{\linewidth}{>{\raggedright\arraybackslash}p{0.16\linewidth} >{\raggedright\arraybackslash}X >{\centering\arraybackslash}p{0.08\linewidth}}
    \toprule
    Model & Answer & Judge \\
    \midrule
    Baseline & "I don't have specific information about your music streaming service." & 2 \\
    Config B & "You have been using Spotify lately to listen to indie and alternative rock music." & 5 \\
    \bottomrule
  \end{tabularx}
\end{table}
Config B's mixed training, which includes LongMemEval examples with user-preference questions, helps the agent locate and retrieve the specific fact from the memory bank.
\paragraph{Example 2: Temporal reasoning (Config C strongest)}
Question: "How many days passed between the day I started watering my herb garden and the day I harvested my first batch of fresh herbs?" Gold answer: "24 days"
\begin{table}[htbp]
  \centering
  \small
  \caption{Model answers for Example 2: Temporal reasoning (Config C strongest).}
  \begin{tabularx}{\linewidth}{>{\raggedright\arraybackslash}p{0.16\linewidth} >{\raggedright\arraybackslash}X >{\centering\arraybackslash}p{0.08\linewidth}}
    \toprule
    Model & Answer & Judge \\
    \midrule
    Baseline & "The exact number of days cannot be determined from the given information." & 3 \\
    Config C & "Based on the provided memories, approximately 33 days passed." & 2 \\
    \bottomrule
  \end{tabularx}
\end{table}
While Config C's answer is numerically incorrect, it attempts temporal reasoning rather than declining to answer --- a behavioral difference attributable to LongMemEval's temporal-reasoning training examples.
\paragraph{Example 3: Single-session retrieval (all trained models improve)}
Question: "What size is my new Samsung TV?" Gold answer: "55-inch"
\begin{table}[htbp]
  \centering
  \small
  \caption{Model answers for Example 3: Single-session retrieval (all trained models improve).}
  \begin{tabularx}{\linewidth}{>{\raggedright\arraybackslash}p{0.16\linewidth} >{\raggedright\arraybackslash}X >{\centering\arraybackslash}p{0.08\linewidth}}
    \toprule
    Model & Answer & Judge \\
    \midrule
    Baseline & "I don't have the specific size of your new Samsung TV." & 4 \\
    Config A & "Your new Samsung TV is 55 inches." & 5 \\
    \bottomrule
  \end{tabularx}
\end{table}
RL training broadly improves the agent's ability to locate specific facts in the memory bank, regardless of curriculum composition.
\end{document}